\pdfoutput=1

\documentclass[11pt]{article}

\usepackage{acl}

\usepackage{times}
\usepackage{latexsym}

\usepackage[T1]{fontenc}

\usepackage[utf8]{inputenc}

\usepackage{microtype}

\usepackage{amsmath}
\usepackage{graphicx}
\usepackage{pifont}
\usepackage{soul}
\usepackage{color}

\definecolor{aquamarine}{rgb}{0.5, 1.0, 0.83}
\definecolor{carnationpink}{rgb}{1.0, 0.65, 0.79}

\DeclareRobustCommand{\hlgreen}[1]{{\sethlcolor{aquamarine}\hl{#1}}}

\DeclareRobustCommand{\hlred}[1]{{\sethlcolor{carnationpink}\hl{#1}}}

\title{Predicting Intervention Approval in Clinical Trials \\ through Multi-Document Summarization}

\author{Georgios Katsimpras \and Georgios Paliouras \\
  NCSR Demokritos, Athens, Greece \\
  \texttt{\{gkatsibras,paliourg\}@iit.demokritos.gr} \\}

\begin{document}
\maketitle
\begin{abstract}

Clinical trials offer a fundamental opportunity to discover new treatments and advance the medical knowledge. 
However, the uncertainty of the outcome of a trial can lead to unforeseen costs and setbacks. In this study, we propose a new method to predict the effectiveness of an intervention in a clinical trial. Our method relies on generating an informative summary from multiple documents available in the literature about the intervention under study. Specifically, our method first gathers all the abstracts of PubMed articles related to the intervention. Then, an evidence sentence, which conveys information about the effectiveness of the intervention, is extracted automatically from each abstract. Based on the set of evidence sentences extracted from the abstracts, a short summary about the intervention is constructed. Finally, the produced summaries are used to train a BERT-based classifier, in order to infer the effectiveness of an intervention. 
To evaluate our proposed method, we introduce a new dataset which is a collection of clinical trials together with their associated PubMed articles.
Our experiments demonstrate the effectiveness of producing short informative summaries and using them to predict the effectiveness of an intervention. 
\end{abstract}

\section{Introduction}
Clinical Trials (CT) present the basic evidence-based clinical research tool for assessing the effectiveness of health interventions. Nevertheless, only a small number of interventions make it successfully through the process of clinical testing. Approximately, 39\%-64\% of interventions actually advance to the next step of each phase of clinical trials \cite{dimasi2010trends}. The uncertainty of a CT outcome could lead to increased costs, prolonged drug development and ineffective treatment for the participants.
At the same time, the volume of published scientific literature is rapidly growing and offers the opportunity to explore a valuable knowledge. Therefore, there is a need to develop new tools which can i) integrate such information and ii) enhance the process of intervention approval in CT.

Predicting the approval of an intervention, a task that describes the ability of a system to predict whether an intervention will reach the final stage of clinical testing, is a topic that has been studied before \cite{gayvert2016data, lo2018machine}. The majority of these studies use various traditional machine learning methods and rely on structured data from various sources, including biomedical, chemical or drug databases \cite{munos2020improving, heinemann2016reflection}. However, only a few studies take into account the textual information that is available online, and mostly in a supplementary manner \cite{follett2019quantifying, geletta2019latent}. In fact, employing natural language processing (NLP) techniques to address the outcome prediction task has been hardly explored. 

Recognising this lack of related studies, the work presented here addresses the task of predicting intervention approval with the use of NLP. Particularly, we relied on generating concise and informative summaries from multiple texts that are relevant to the intervention under evaluation. In a sense, we built an intervention-specific narrative which combines key information from multiple inter-connected documents. The benefit of using multiple articles to generate summaries is that they can cover the inherently multi-faceted nature of an intervention's clinical background. 

More precisely, given an intervention, our system retrieves all PubMed abstracts that are relevant to the intervention and refer to a clinical study. It then extracts the evidence sentences from each abstract using a BERT-based evidence sentence classifier, in a similar fashion to \cite{deyoung2020evidence}. This set of evidence sentences, which captures the consolidated narrative about the intervention, can grow gradually, as new articles become available. Thus, further analysis is necessary in order to select the most important information. Using the set of evidence sentences for each intervention, we generate short summaries by leveraging the power of language models (BERT or BART). The resulted summaries are then fed to a BERT-based binary sequence classifier which makes a prediction about the likely approval or not of the intervention.

Overall, the main contributions of the paper are the following:
\begin{itemize}

    \item We propose a new approach for predicting the approval of an intervention which is based on a three-step NLP pipeline.
    \item We provide a new dataset for the task of intervention approval prediction that consists of 704 interventions and 15,800 PubMed articles in total. 
    \item We confirm through experimentation the effectiveness of the proposed approach.
\end{itemize}

\section{Related Work}

\paragraph{} 
\textbf{Intervention Success Prediction} The prediction of intervention approval belongs to a broader category of medical prediction tasks. Relevant work includes clinical trial outcome prediction \cite{munos2020improving, tong2019machine, hong2020predicting},  drug approval \cite{gayvert2016data,lo2018machine, siah2021predicting, heinemann2016reflection}, clinical trial termination \cite{follett2019quantifying, geletta2019latent, elkin2021predictive}, predicting phase transition \cite{hegge2020predicting, qi2019predicting}. All these studies rely either on specific types of structured data or on combining structured data with limited unstructured data.

Differently from this line of work, the authors of \cite{lehman2019inferring} proposed an approach that employs NLP to infer the relation between an intervention and the outcome of a specific clinical trial. Their method is based on extracting evidence sentences from unstructured text. An extension of this work suggests the use of BERT-based language models for the same task \cite{deyoung2020evidence}. Another closely related study \cite{jin2020predicting}, performs a large-scale pre-training on unstructured text data to infer the outcome of a clinical trial. Our approach builds upon this related work, aiming to incorporate information from multiple articles. This extension is motivated by the assumption that the inter-connected clinical knowledge, coming from multiple sources can provide a more holistic picture of the intervention, facilitating more precise analysis and accurate prediction.   

Although all these prior efforts tackle, more or less, the problem of intervention approval, none of them attempted to predict the effectiveness of an intervention using summarization methods.

\textbf{Summarization} The goal of summarization is to produce a concise and informative summary of a given text. There are two main categories of approaches: i) \emph{extractive}, which tackles summarization by selecting the most salient sentences from the text without changing them, and ii) \emph{abstractive}, which attempts to generate out-of-text words or phrases instead of extracting existing sentences. Early systems were primarily extractive and relied on sentence scoring, selection and ranking \cite{allahyari2017text}. However, both extractive and abstractive approaches have advanced significantly due to the novel neural network architectures, such as Transformers \cite{vaswani2017attention}. The Transformers architecture is utilized by the BERT \cite{devlin2018bert} and BART \cite{lewis2019bart} language models which are used by the state-of-the art solutions for multiple NLP tasks, including summarization. Although most of the summarization literature focuses on single-document approaches, there is also a line of work that applies summarization on a set of documents, i.e. multi-document summarization \cite{ma2020multi}.  Such approaches are of particular relevance to our work, as we aim to summarize a set of sentences about a particular intervention.

\textbf{Summarization in the Medical Domain}
Summarization has been used to address various problems in the field of medicine. These include electronic health record summarization \cite{liang2019novel}, medical report generation \cite{zhang2019optimizing, liu2021competence}, medical facts generation \cite{wallace2021generating, wadden2020fact} and medical question answering \cite{demner2006answer, nentidis2021overview}. 

Our work is inspired by recent work on multi-document summarization of medical studies \cite{deyoung2021ms2}. Apart from introducing a new summarization dataset of medical articles, that work also proposed a method to generate abstractive summaries from multiple documents. Their model is based on the BART language model, appropriately modified to handle multiple texts. Our model differs in the way it handles the input texts. Instead of concatenating all texts into a single representative document, we order them chronologically and split them into equal-size chunks. Doing so, we expect the clinical studies that were conducted during a similar time period, to reside in the same chunk.

\section{Task Overview}

\begin{figure}[htp]
\begin{center}
\includegraphics[width=\linewidth]{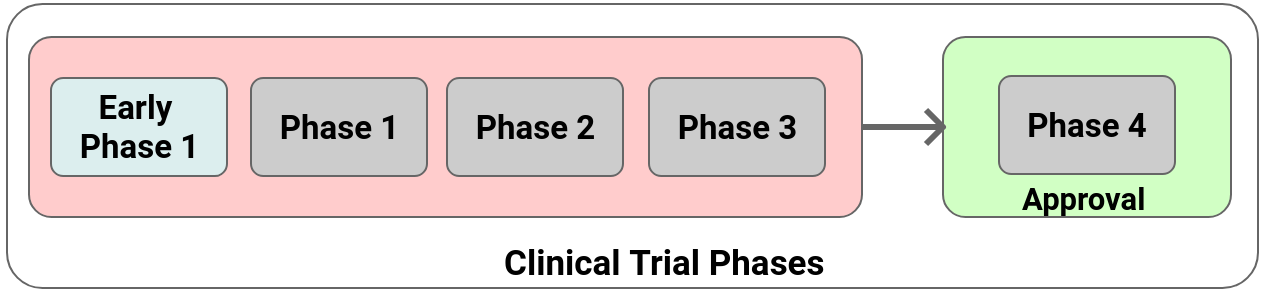}
\caption{The phases of a clinical trial.}
\label{fig:ct_phases}
\end{center}
\end{figure}

According to the U.S. Food and Drug Administration (FDA), a CT addresses one of five phases of clinical assessment: Early Phase 1 (former Phase 0), Phase 1, Phase 2, Phase 3 and Phase 4. Each phase is defined by the study's objective, the interventions under evaluation, the number of participants, and other characteristics\footnote{https://clinicaltrials.gov/}. Notably, Phase 4 clinical trials take place after FDA has \emph{approved} a drug for marketing. Therefore, we can assume that a CT in Phase 4 assesses effective intervention. On this basis, our task is to predict whether an intervention will advance to the final stage of clinical testing (Phase 4), as shown in Figure \ref{fig:ct_phases}. 

We model the task of predicting the success or failure of an intervention as a binary classification task. 
All data relevant to Phase 4 are omitted from the training stage. 

\section{Data}
In this work, we introduce a new dataset\footnote{https://github.com/nneinn/ct\_intervention\_approval} for the task of predicting intervention approval. The dataset is a collection of structured and unstructured data in English derived from clinicaltrials.gov and PubMed during May-June 2021.

As a first step in the construction of the dataset, we retrieve all available CT studies from clinicaltrials.gov that satisfy some criteria. Then, we associate each CT with PubMed articles based on the CT study identifier. Following some cleaning process (i.e. deduplication and entity resolution) we generate the final dataset.

\textbf{Clinical Trials Studies}
At the time of writing, more than 350,000 studies were available online at \emph{clinicaltrials.gov}.  We focused on \emph{cancer} related clinical testing and we retrieved approximately 85,000 studies related to this topic using a list of associated keywords\footnote{The complete list of the keywords used is: cancer, neoplasm, tumor, oncology, malignancy, neoplasia, neoplastic syndrome, neoplastic disease,	neoplastic growth and malignant growth}. 

From this set, we were interested in interventional clinical trials and specifically in two categories that indicate the status of the trial: i) \emph{``Completed''}, meaning that the trial has ended normally, and ii) \emph{``Terminated''}, meaning that the trial has stopped early and will not start again. The resulting set of studies contains 34,517 completed and 6,872 terminated trials. 

\textbf{Interventions Dataset}
Using the selected CTs, we associated each intervention with its corresponding trials. Therefore, a clinical trial record was formed for each intervention. Then, we selected all interventions that are assessed in at least one Phase 4 CT to form our positive target class (i.e. approval). Likewise, we built our negative target class (i.e. termination) using interventions that led to a trial termination. In total, our dataset contains 404 approved and 300 terminated interventions.

\begin{figure*}[ht!]
    \centering
    \includegraphics[width=\textwidth]{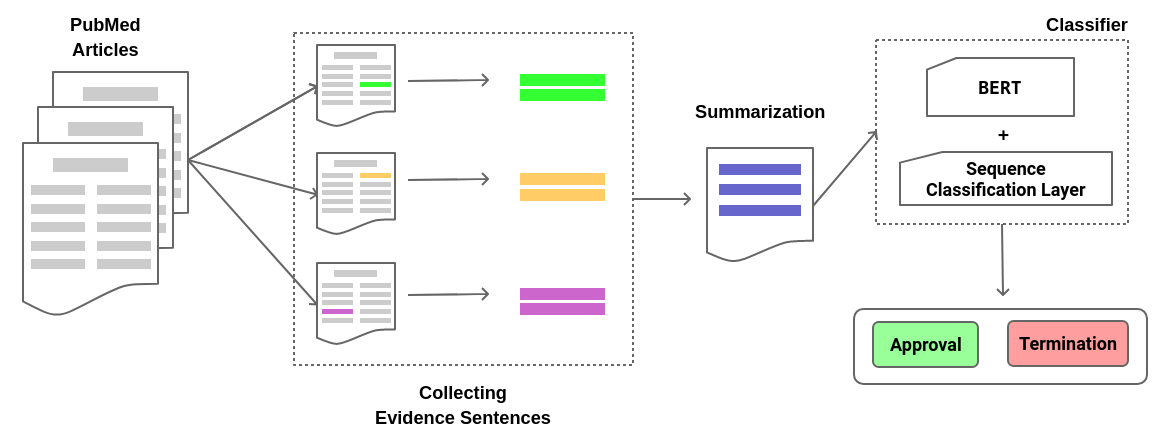}
    \caption{Overview of the proposed approach for classifying an intervention.}
    \label{fig:flow}
\end{figure*}

For each intervention, we collect all articles from PubMed that are explicitly related to one of the CTs of the intervention. To achieve this, we combine two approaches. First, we search for eligible articles (or links to articles) in the corresponding structured results of \emph{clinicaltrials.gov}. Secondly, we use the CT unique identifiers to query the PubMed database. Then, the selected PubMed articles are associated with the intervention. This way an intervention is linked with multiple studies that are inter-connected, and thus an intervention-specific narrative is developed. In our dataset, an intervention is associated on average with 22.4 pubmed articles, though for terminated interventions this number is just 1.4. This is because terminated interventions are usually not assessed in many CTs. Overall, our dataset contains 15,800 pubmed articles. The details of the dataset are presented in Table \ref{tab:dataset}. 

\begin{table}[htp]
\centering
\begin{tabular}{lccccc}
\hline
\textbf{Type} &\textbf{|I|} & \textbf{|A|} & \textbf{avg} \\
\hline

Approved & 404 & 15,379 & 38.1\\ 
Terminated & 300 & 421 & 1.4\\ 
\hline
\textbf{Total} & 704 & 15,800 & 22.4\\ 

\hline
\end{tabular}

\caption{The details of the interventions dataset. |I|, |A| and \emph{avg} denote the number of interventions, the number of articles and the average number of articles per intervention respectively.}
\label{tab:dataset}
\end{table}

In addition, we attempted to evaluate\footnote{The results on this dataset are presented in appendix \ref{sec:appendix}} our approach on a previously used dataset \cite{gayvert2016data}, which consists of 884 (784 approved, 100 terminated) drugs along with a set of 43 features, including molecular properties, target-based properties and drug-likeness scores. 

\section{Methodology}

In Figure \ref{fig:flow}, we illustrate the proposed approach, which consists of three main steps. Initially, we use the abstracts of the intervention's clinical trial record to extract evidence sentences. These sentences are then used to generate a short summary that contains information about the efficacy of the intervention. The summary is then processed by a BERT-based sequence classifier to make the final decision about the intervention.  Each of the three steps is detailed in the following subsections.

\subsection{Evidence Sentences}
\label{sec:evid_sent}
\begin{figure}[ht!]
    \centering
    \includegraphics[width=0.48\textwidth]{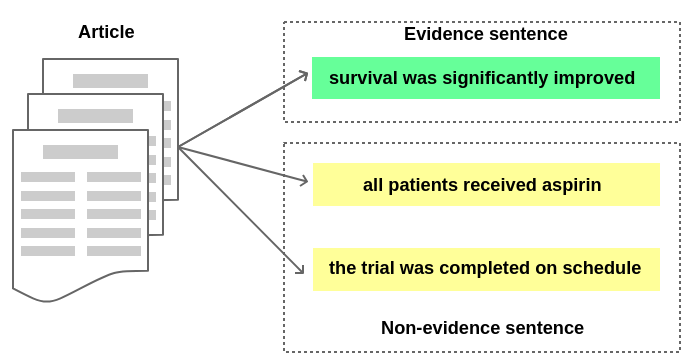}
    \caption{Evidence sentence identification. The evidence sentences constitute the positive instances whereas the non-evidence sentences the negative ones.}
    \label{fig:ev_example}
\end{figure}

Identifying evidence bearing sentences in an article for a given intervention is an essential step in our approach. Differently from other sentences in an article, evidence sentences contain information about the effectiveness of the intervention (Figure \ref{fig:ev_example}). Therefore, it is crucial that our model has the ability to discriminate between evidence and non-evidence sentences.

First, all abstracts related to the given intervention are broken into sentences. The sentences of each abstract are then processed one-by-one by a BERT-based classifier that estimates the probability of each sentences containing evidence about the effectiveness of the intervention. For the classifier, we selected a version of the PubMedBERT \cite{gu2020domain} model, which is pre-trained only on abstracts from PubMed.  We tested several models, including BioBERT \cite{lee2020biobert}, clinicalBERT \cite{alsentzer2019publicly} and RoBERTa \cite{liu2019roberta}, but PubMedBERT performed the best in our task. On top of PubMedBERT, we trained a linear classification layer, followed by a Softmax, using the dataset from \cite{deyoung2020evidence}. This dataset is a corpus especially curated for the task of evidence extraction and consists of more than 10,000 annotations. The classifier is trained with annotated evidence sentences (i.e. positive samples) and a random sample of non-evidence sentences (i.e. negative samples). Regarding the ratio of positive to negative samples, cross-validation on the training set showed 1:4 to be a reasonable choice. The evaluation of the different BERT-based models was done based on the same data splits (train, test and validation) as in \cite{deyoung2020evidence}.

Once scored by the classifier, the highest scoring sentence is selected from each abstract. Therefore, for each intervention we extract as many sentences as the number of abstracts in its clinical record.

\subsection{Short Summaries}
To generate short and informative summaries we explore both extractive and abstractive approaches.

\textbf{Extractive}
Summaries were based on the evidence sentences extracted in the previous step. Specifically, we re-rank them and choose the top $k$ $(k=5)$ to compose our final summary. The model we use here is the same BERT-based model as in Section \ref{sec:evid_sent}.

\textbf{Abstractive}
Considering that an intervention is linked to multiple abstracts and thus to multiple evidence sentences, we first order all evidence sentences chronologically and combine them into a single text. Then, we split them to equal chunks\footnote{A chunk has length equal to the maximum input length of the BART model (1024).} and each chunk then is fed to a BART-based model to produce the final summary.

BART has been shown to lead to state-of-the-art performance on multiple datasets \cite{fabbri2021summeval}. Specifically, we used the pre-trained distilBART-cnn-12-6 model which is trained on the CNN summarization corpus \cite{cnntext}. Since abstractive summarization produces out-of-text phrases, it needs to be fine-tuned with domain knowledge. In our case, we fine-tuned the BART model with the MS2 dataset \cite{deyoung2021ms2}, which contains more than 470K articles and 20K summaries of medical studies. 
\paragraph{}
We limited the length of the output summary to 140 words.  For the extractive setting, in case the top $k$ sentences exceeded this limit, we removed the extra words. For the abstractive setting we iteratively summarized and concatenated the chunks for each intervention until the expected number of 140 words was accomplished.

\subsection{Inferring Efficacy}
We model the task of inferring the approval of an intervention as a binary classification task. In our approach, each intervention is represented by a short summary. For the classification of the summaries, we used again a PubMedBERT model. On top of it, we trained a linear classification layer, followed by a sigmoid, using the summaries generated in the previous step: Our positive training instances were the summaries of interventions that have been approved, and correspondingly, the negative ones were the summaries of interventions that have been terminated. Hence, the model decides on the approval of the interventions.  

\subsection{Technical set-up}
All models were pre-trained and fine-tuned for the corresponding task. The maximum sequence size was 512 and 1024 for BERT-based and BART-based models respectively. The Adam optimizer \cite{kingma2014adam} was used to minimize the cross-entropy losses with learning rate 2e-5 and epsilon value 1e-8 for all models. We trained all models for 5 epochs, with batch sizes of 32, except the abstractive summarizer for which the batch size was decreased to 4 due to RAM memory limitations of our system. The implementation was done using the HuggingFace library \cite{wolf2020transformers} and Pytorch\cite{NEURIPS2019_9015}.

\section{Results and Analysis}
We followed different training approaches for the different trainable components of our pipeline. For the evidence sentence selection and the abstractive summarization models we split the data into development and test and then split the development set further into training (90\%) and validation (10\%). We kept the model  that performed best on the validation set and evaluated it on the held-out test set of each task respectively, averaged over three random data splits. Considering the small size of the interventions dataset, we applied a 10-fold cross validation for the final classification task. For this task, we report macro averages of the evaluation metrics over the ten folds.

\subsection{Ablation Study}
Our experimentation started with a comparison of different variants and choices that were available for the various modules of our approach.

\textbf{Evidence Classifier}
Coming early in the pipeline, the performance of the evidence classifier can play a significant role in downstream tasks. The chosen approach relied on domain-specific BERT models. As domain-specific training that can affect the performance of BERT-based models, we conducted a comparison between different variants of BERT. The results in Table \ref{tab:ev_s} demonstrate that the performance of the models is comparable, with all models obtaining scores over 90\% in terms of F1 and AUC. PubMedBERT model achieved the best scores and was used in the rest of the experiments.

\begin{table}[thp]
\centering
\begin{tabular}{lcccc}
\hline
\textbf{Model} & \textbf{P}& \textbf{R} & \textbf{F1} &\textbf{AUC}\\
\hline

BioBERT & .928& .938 & .933 & .957\\ 
ClinicalBERT  & .913& .925& .919 & .945\\ 
RoBERTa  & .905& .919& .912 & .931\\ 
PubMedBERT  & \textbf{.931}& \textbf{.956} & \textbf{.943} & \textbf{.969}\\ 

\hline

\hline
\end{tabular}

\caption{The results of the domain-specific BERT variants that were used for the evidence classifier. All models were trained with negative sampling ratio 1:4. The results denote the averages over three random train-test splits.}
\label{tab:ev_s}
\end{table}

\textbf{Summarization Adequacy}
\label{sec:summ}
We assess the performance of the summarization methods on the MS2 dataset which is a collection of summaries extracted from medical studies. The task of the summarizers is to produce texts that approximate the 
target summaries. 
We measure the performance of the summarization methods using ROUGE and the results are presented in Table \ref{tab:sum_a}. As expected, the abstractive method achieves higher scores, as it has more flexibility in forming summaries. We also observed that domain-specific training improves performance. The $abstractive_{no}$ model is a generic BART model without fine-tuning in the domain. Comparing its performance to the $abstractive$ model, which was fine-tuned on a small sample of the MS2 dataset that was excluded from the evaluation process, we notice a statistically significant improvement.

\begin{table}[htp]
\centering
\begin{tabular}{lccc}
\hline
\textbf{Model} &\textbf{R-1} & \textbf{R-2} & \textbf{R-L}\\
\hline

abstractive$_{no}$ & 24.85 & 4.34 & 15.48\\ 

abstractive & \textbf{39.38} & \textbf{11.98} & \textbf{20.13}\\ 

extractive & 19.24 & 3.22 & 13.19\\ 
\hline

\hline
\end{tabular}

\caption{Evaluation of summarization methods on the MS2 dataset. The abstractive$_{no}$ refers to the generic BART model without any fine-tuning in the domain.}
\label{tab:sum_a}
\end{table}

Abstractive methods seem to provide better summaries, however, whether these are more useful than the extractive summaries for our donwstream task is still to be determined.

\subsection{Predicting Intervention Efficiency}

Having made the choices for the individual modules, we now turn to the ultimate task, which is the prediction of the efficiency of the intervention. We evaluate two variations of our proposed method; i) with abstractive summarization denoted as PIAS$_{abs}$ and ii)  with extractive summarization denoted as PIAS$_{ext}$. We compare their performance against two baselines:

\begin{itemize}
    \item \textbf{BS}: This is a PubMedBERT model that is trained with a single evidence sentence per intervention (instead of a summary). The sentence is extracted from the most recent PubMed article relevant to the intervention.
    \item \textbf{BN}: This is similar to $BS$ but instead of using a single sentence for each intervention it is trained with $n$ evidence sentences extracted from $n$ different articles $(n=3)$.  The articles are selected randomly among the ones referring to the intervention.

\end{itemize}

The performance of all models is shown in Table \ref{tab:main}. The proposed method outperforms the baselines independent of the summarization methods that is used. Interestingly, even selecting randomly selected  evidence sentences seem to help, as $BN$ achieved a higher performance than $BS$. Still, the use of summarization provides a significant boost over both baseline methods, validating the value of using short summaries to evaluate the efficiency of an intervention. Models that do not take advantage of the inter-connected documents suffer a significant drop in performance. Thus, this result justifies the design of the proposed method.

We can also observe that the best performance of the proposed method is achieved when using the extractive summarization method. Extractive summaries have demonstrated low ROUGE scores in Section \ref{sec:summ}. Still, they can properly capture the properties involved in the data for the classification task. On the other hand, although the abstractive summarizer achieved better ROUGE  scores, it seems that the generated summaries cannot discriminate the target classes (approved or terminated) as well as the extractive ones. This indicates that the quality of the summary, in terms of the ROUGE score, is not decisive in the classification of the intervention.

\begin{table}[htp]
\centering
\begin{tabular}{lcccc}
\hline
\textbf{Model} &\textbf{P} & \textbf{R} & \textbf{F1}\\
\hline
BS & .717 & .706 & .702 \\ 
BN & .732 & .731 & .731 \\ 
PIAS$_{abs}$ & .781 & .774 & .773 \\ 

PIAS$_{ext}$ & \textbf{.796} & \textbf{.793} & \textbf{.792}\\ 
\hline
\end{tabular}

\caption{The classification results of all models. The reported precision, recall and f1 scores the macro averages over ten folds.}
\label{tab:main}
\end{table}

Analyzing further the performance of our best model, PIAS$_{ext}$, we report macro average scores for each target class in Table \ref{tab:class}.
\begin{table}[htp]
\centering
\begin{tabular}{lcccc}
\hline
\textbf{class} &\textbf{P} & \textbf{R} & \textbf{F1}\\
\hline
positive (approved) & .808 & .819 & .815 \\ 
negative (terminated) & .778 & .765 & .772 \\ 
\hline
\end{tabular}

\caption{The performance of our best model, i.e. PIAS$_{ext}$,  for each target class. The scores denote macro averages over ten folds.}
\label{tab:class}
\end{table}
We notice that the model is slightly better at predicting the approval of an intervention rather than its termination. 
This can be explained by the fact that the approved interventions are associated with a considerably larger number of articles than the terminated ones. This leads to richer summaries for the approved interventions and thus to a more informed decision.

\subsection{Predicting Phase Transition}

\textbf{Early prediction of approval} To build our models, we considered all the available data from Phase 1, Phase 2 and Phase 3. However, predicting the success of an intervention at the earliest phase possible is compelling. Therefore, we examine the ability of our model in making early predictions. More precisely, we evaluate PIAS$_{ext}$ model on the following three transitions: Phase 1 to Approval, Phase 2 to Approval and Phase 3 to Approval. 

To perform this experiment, we select the interventions that have CTs in various stages and there is least one article for each phase. 
In total, this subset contains 249 interventions (193 approved and 56 terminated). Then, we use 80\% for  training and 20\% for testing. For each transition, we train our model only with training instances from the corresponding phase. In Table \ref{tab:phase1}, we report the macro average scores over ten random splits of the data.  

\begin{table}[htp]
\centering
\begin{tabular}{lcccc}
\hline
\textbf{transition} &\textbf{P} & \textbf{R} & \textbf{F1}\\
\hline
phase1$\xrightarrow{to}$approval & .39 & .50 & .44 \\ 
phase2$\xrightarrow{to}$approval & .78 & .70 & .72 \\ 
phase3$\xrightarrow{to}$approval & .81 & .84 & .82 \\ 

\hline
\end{tabular}

\caption{The performance of our best model, i.e. PIAS$_{ext}$, in predicting phase-to-approval transitions. The scores denote the averages over ten random runs.}
\label{tab:phase1}
\end{table}

The results indicate that prediction of approval, while at Phase 1 is very hard, but the transition from Phase 2  and Phase 3 to approval can be predicted with considerable success. The large gap in performance between Phase 1 and Phase 2, 3 transitions is explained by the lack of clinical evidence in early phases.

\textbf{Phase to Phase}
Another interesting and challenging task is to predict the transition of an intervention to the next phase of the clinical trial process. 
In this experiment, we want to predict Phase 1 to Phase 2 and Phase 2 to Phase 3 transitions. For each transition, we use data only from the former phase for training (e.g. for Phase 2 to Phase 3 transition we use data from Phase 2) for both target classes. Again, we use 80\% for training and 20\% for testing and present the average scores over ten random splits.

\begin{table}[htp]
\centering
\begin{tabular}{lcccc}
\hline
\textbf{transition} &\textbf{P} & \textbf{R} & \textbf{F1}\\
\hline
phase2$\xrightarrow{to}$phase3 & .84 & .82 & .83 \\ 
phase1$\xrightarrow{to}$phase2 & .77 & .76 & .77 \\ 
\hline
\end{tabular}

\caption{The performance of our best model, i.e. PIAS$_{ext}$, in predicting phase-to-phase transitions. The scores denote the averages over ten random runs.}
\label{tab:phase2}
\end{table}

Table \ref{tab:phase2} shows the results for the two transitions, which are comparable to the overall predictive performance of the model.
Considering the small size of the datasets used in both phase transition tasks, these results can serve only as an indication of how our model behaves. Further analysis and experiments should be conducted for a more thorough evaluation.

\subsection{Explainability of Predictions}
It is clinically very valuable to identify the factors that contribute most to a particular decision of the classifier.
Interestingly, the summaries generated from our models can also serve that purpose very well.

\begin{table*}[htb]
\centering
\begin{tabular}{p{0.12\linewidth}|c|p{0.37\textwidth}|p{0.37\textwidth}}
\hline
\textbf{Intervention} &\textbf{} &\textbf{PIAS$_{abs}$} & \textbf{PIAS$_{ext}$}\\
\hline

pertuzumab & \ding{51}  & \small{the primary endpoint of the study is progression-free survival.  median progression- \hlgreen{free survival was 12.4 months in the control group, as compared with 18.5months in the pertuzumab group.}  median survival was  <dig> months, 12.3 months, and 12.5 months, respectively, in the p=0·0141 group and p =0·0\% in the qtl group, respectively.  the p <dig) group \hlgreen{had a significantly improved pathological complete response rate} compared with the group without complete response.  p=dig> month and qtl were the most significantly different groups in both groups.  p =dig>  }
& \small{\hlgreen{Disease-free survival results were consistent with progression-free survival results and were 81\% (95\% CI 72-88) for group A, 84\% (72-91) for group B, 80\% (70-86) for group C, and 75\% (64-83) for group D}. Patients who achieved total pathological complete Three patients [1.5\%; 95\% confidence interval (CI) 0.31\% to 4.34\%] in cohort A experienced four New York \hlgreen{No evidence of DDIs for pertuzumab on trastuzumab, trastuzumab on pertuzumab, or pertuzumab on chemotherapy PK was observed.} The median progression-free survival (PFS) among patients who received NAT was 15.8 months compared with CNS ORR was 11\% (95\% CI, 3 to 25), with four partial responses (median duration of response, 4.6 months).} \\ \hline
taxane & \ding{55} & \small{the most common \hlred{serious adverse events were} anaemia, upper gastrointestinal haemorrhage, pneumonia, and pneumonia in the trastuzumab emtansine 24 mg/kg weekly group compared with \hlred{pneumonia, febrile neutropenia, and anaemia in the taxane group. median overall survival was 11.8 months with trastzumab 2.4 mg/ kg weekly and 10.0 months with taxane.2) with taxanes}.3) with t-dm1 was not associated with superior os or superior os versus taxane in any subgroup.5–10\% of the patients with high body weight and low baseline trast}
& \small{The most common serious adverse events were anaemia (eight 4), upper gastrointestinal haemorrhage (eight 4), pneumonia (seven 3), gastric haemorrhage (six 3), and gastrointestinal haemorrhage (five 2) in the trastuzumab emtansine 24 mg/kg weekly group compared with \hlred{pneumonia (four 4), febrile neutropenia (four 4), anaemia (three 3), and neutropenia (three 3) in the taxane group}. Median overall survival was 11.8 months (95 confidence interval ci, 9.3-16.3) with trastuzumab emtansine 2.4 mg/kg weekly \hlred{and 10.0 months (95 ci, 7.1-18.2) with taxane (unstratified hazard ratio 0.94, 95 ci, 0.52-1.72)}}.\\
\end{tabular}
\caption{Examples of generated summaries from our models. These summaries can be used to explain the predictions of the classifier. The second column displays the prediction of the classifier for the specific intervention;  \ding{51} denotes approval and \ding{55} denotes termination. }
\label{tab:summaries}
\end{table*}
\vfill

Table \ref{tab:summaries} illustrates some examples of interventions along with their abstractive and extractive summaries as produced by our pipeline. For the first intervention, \emph{pertuzumab}, it is notable that both summaries report a improved median progression-free survival which somewhat explains the prediction. For the second intervention, \emph{taxane}, the summaries mention the greater incidence of serious adverse events and lower median overall survival, which counts against the approval of the intervention. We also notice that many numerical entities are randomly placed or changed in the abstractive summary. This contributes to  the tendency of the abstractive methods to generate "hallucinated" evidence, as observed in the literature \cite{cao2018faithful}. However, the abstractive summaries look more readable. A more exhaustive analysis, including also a human evaluation, is needed to assess the ultimate explainability of these summaries. 

\section{Conclusion}
Predicting intervention approval in clinical trials is a major challenge with significant impact in healthcare. In this paper, we have proposed a new pipeline to address this problem, based on state-of-the-art NLP techniques. The proposed method consists of three steps. First, it identifies evidence sentences from multiple abstracts  related to an intervention. Then, these sentences are used to produce short summaries. Finally, a classifier is trained on the generated summaries in order to predict the approval or not of an intervention.

Moreover, we introduced a new dataset for this task which contains 704 interventions associated with 15,800 abstracts. This data was used to evaluate our pipeline against other baseline models. The experimental results verified the effectiveness of our approach in predicting the approval of an intervention and the contribution of each step of the proposed pipeline to the final result. Further evaluation on predicting phase transitions, showed that our model can assist in all stages of a clinical trial. Besides, the generated multi-document summaries can be naturally used to explain the predictions of the model.

There are multiple ways to extend this work. In terms of multi-document summarization, there is room to explore more advanced summarization models, quality and performance metrics as well as better explainability assessment. In the bigger picture, we shall also consider to expand the dataset by extending its size and incorporating different types of resources (e.g. drug interaction networks). Finally, we are interested in enhancing the proposed method to incorporate temporal information associated with the CTs to maintain the history of clinical changes.

\section*{Acknowledgements}
We would like to thank the anonymous reviewers for their valuable and constructive comments on this research.
This works was partially supported by the ERA PerMed project P4-LUCAT (Personalized Medicine for Lung Cancer Treatment:Using Big Data-Driven Approaches For Decision Support) ERAPERMED2019-163. 

\bibliography{acl_latex}
\bibliographystyle{acl_natbib}

\newpage
\appendix

\section{Results on Proctor Dataset}
\label{sec:appendix}
To further evaluate our method, we attempted a comparison with the method presented in \cite{gayvert2016data} using their data. The data contains a list of approved and terminated drugs together with various features. Using this dataset, we experienced two issues that made the comparison incomparable: i) For many drugs we could not find relevant articles in PubMed. The original dataset contains 828 drugs whereas we managed to collect information only for 537. Thus, the scores of our method are not directly comparable to the ones reported in \cite{gayvert2016data} ii) Four important features that were used in \cite{gayvert2016data} are missing in the dataset. Therefore, the reproduction of the exact model is not possible.

Despite these facts, we performed a comparison of the methods for the subset that we collected:

\begin{itemize}
    \item \textbf{RF$_{1}$}: This model reports the scores from \cite{gayvert2016data}.
    \item \textbf{RF$_{2}$}: This is a Random Forest model similar to the original one, but it is trained only with the available features.

\end{itemize}

The overall performances of all models are depicted in Table \ref{tab:proc}.

\begin{table}[htp]
\centering
\begin{tabular}{lc}
\hline
\textbf{Model}  & \textbf{AUC}\\
\hline
RF$_{1}$ & .826 \\ 
RF$_{2}$  & .484 \\ 

PIAS$_{ext}$  & .586\\ 
\hline
\end{tabular}

\caption{The classification results of all models on the Proctor dataset. The reported precision, recall and f1 scores are macro averages over ten folds.}
\label{tab:proc}
\end{table}
\end{document}